\documentclass[letterpaper, 10 pt, conference]{ieeeconf}  

\IEEEoverridecommandlockouts                              

\usepackage{graphicx}
\usepackage{bm}
\usepackage{amsmath} %
\usepackage{amssymb}  %
\usepackage{multirow}
\usepackage{todonotes}
\setuptodonotes{inline}
\usepackage{url}
\usepackage{algorithm}
\usepackage{algpseudocode}
\usepackage{siunitx}
\usepackage[hidelinks]{hyperref}
\usepackage{cleveref}
\usepackage{comment}
\usepackage{booktabs}
\usepackage{tabularx}
\usepackage{subcaption}

\usepackage[style=ieee, doi=false, isbn=false, url=false]{biblatex} 
\AtBeginBibliography{\small}
\title{\LARGE \bf
Rotograb: Combining Biomimetic Hands with Industrial Grippers using a Rotating Thumb
}

\author{Arnaud Bersier$^{1\dagger}$, Matteo Leonforte$^{1\dagger}$, Alessio Vanetta$^{1\dagger}$, Sarah Lia Andrea Wotke$^{1\dagger}$,\\ Andrea Nappi$^{1}$, Yifan Zhou$^{1}$, Sebastiano Oliani$^{1}$, Alexander M. Kübler$^{1}$, and Robert K. Katzschmann$^{1*}$ 
\thanks{$^{1}$Soft Robotics Lab, IRIS, D-MAVT, ETH Zurich, Switzerland}%
\thanks{$\dagger$ Equal contribution.}
\thanks{$*$ Corresponding author: \href{mailto:rkk@ethz.ch}{\tt\footnotesize rkk@ethz.ch}}
}

\begin{document}
\maketitle
\thispagestyle{empty}
\pagestyle{empty}

\begin{abstract}
The development of robotic grippers and hands for automation aims to emulate human dexterity without sacrificing the efficiency of industrial grippers.
This study introduces Rotograb, a tendon-actuated robotic hand featuring a novel rotating thumb. The aim is to combine the dexterity of human hands with the efficiency of industrial grippers. The rotating thumb enlarges the workspace and allows in-hand manipulation. A novel joint design minimizes movement interference and simplifies kinematics, using a cutout for tendon routing. We integrate teleoperation, using a depth camera for real-time tracking and autonomous manipulation powered by reinforcement learning with proximal policy optimization.
Experimental evaluations demonstrate that Rotograb's rotating thumb greatly improves both operational versatility and workspace. It can handle various grasping and manipulation tasks with objects from the YCB dataset, with particularly good results when rotating objects within its grasp. 
Rotograb represents a notable step towards bridging the capability gap between human hands and industrial grippers. The tendon-routing and thumb-rotating mechanisms allow for a new level of control and dexterity. Integrating teleoperation and autonomous learning underscores Rotograb's adaptability and sophistication, promising substantial advancements in both robotics research and practical applications.

\end{abstract}

\section{Introduction}

\begin{figure}[ht]
        \centering
        \includegraphics[width=\linewidth]{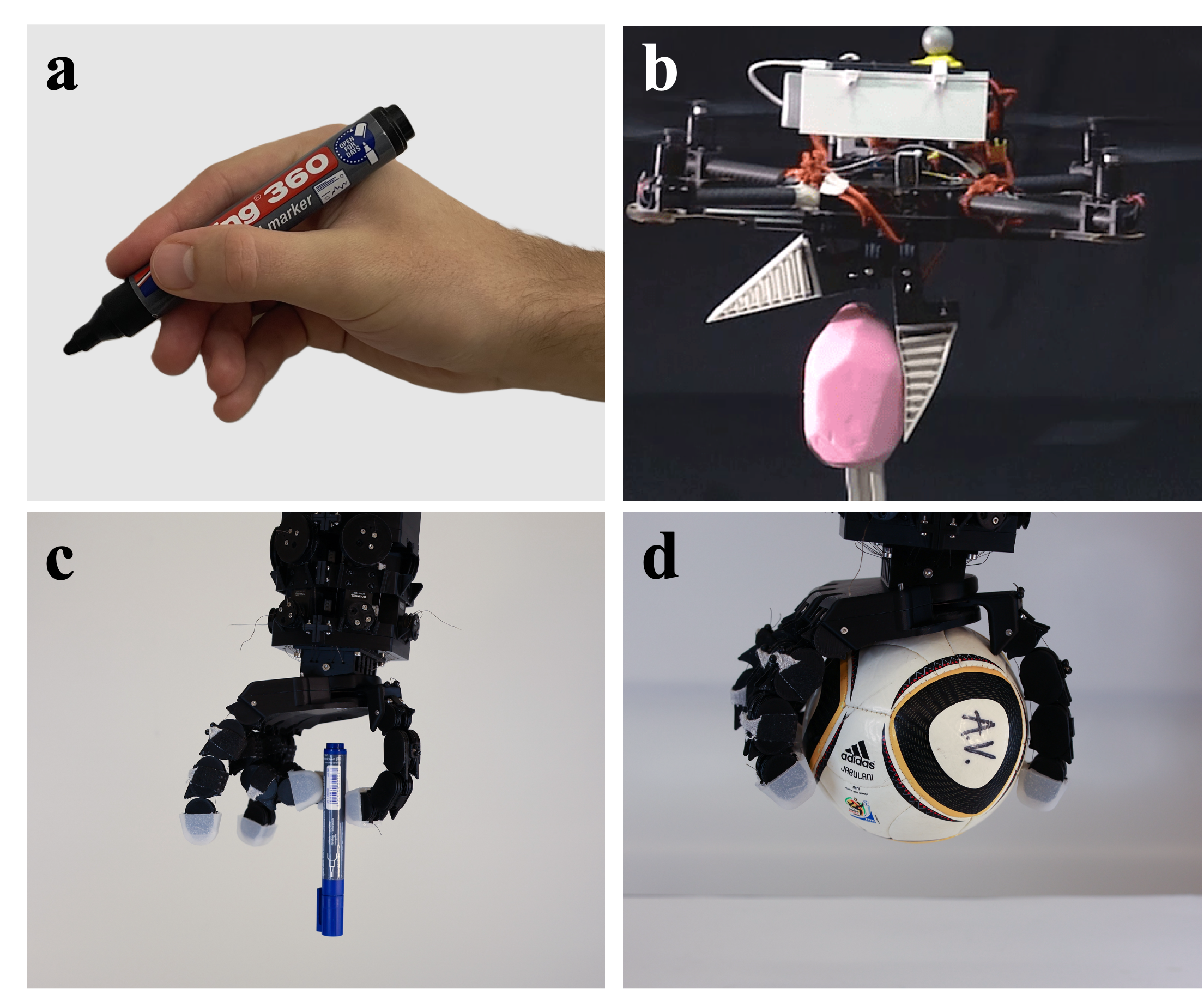}
        \label{fig:hand_cup}
        \caption{The Rotograb robotic gripper \textbf{(c, d)} enables precision and power grasps by integrating a rotating thumb that merges the dexterity of human hands \textbf{(a)} with the power grasp efficiency of soft grippers \textbf{(b)}. Example grasps showing the dexterity of a human hand \textbf{(a)} and the power grasp capability of a soft fin ray gripper \textbf{(b)}~\cite{appius2022raptor}. Our new robotic gripper, Rotograb, uses a rotating thumb to combine the dexterity of robotic hands with industrial grippers to achieve precision grasps \textbf{(c)} and power grasps \textbf{(d)}.}
        
\end{figure}

\subsection{Motivation}

Integrating robots into daily human life requires their interaction with complex surrounding environments. One of the biggest challenges they face is handling objects in a skilled and effective way. This is due to the requirement for multiple degrees of freedom and precise control, which is essential for dealing with a wide variety of objects and tasks.

Developing a gripper for performing a single action can be achieved by optimizing the design for the specific task. However, building a versatile manipulator presents a much more complex challenge. It is tempting to mirror a biomimetic hand, given that the world around us is shaped to be human-friendly. However, this is not necessarily the optimal and most effective approach. For example, a more straightforward single degree-of-freedom gripper designed like a crane may handle grasping large and heavy objects more efficiently.

In this paper, we attempt to challenge the common conception that humanoid grasp is the only possible approach, proposing a hybrid design that combines the powerful grasp of an industrial gripper with a biomimetic design for tasks that require dexterity.

\subsection{Related Work}

The human hand, with its 21 degrees of freedom (DoF), excluding the wrist, is a highly dexterous manipulator that allows a wide range of motions~\cite{rahman2013design}. The thumb alone has five DoF, which makes it opposable by bending the palm to grasp large objects. 
However, replicating such versatility in robotic hands often results in increased complexity and fragility. Nature provides efficient alternatives for strong grasps, like the eagle's grasp, which industrial grippers have emulated for simplicity and strength.

Robotic manipulators are broadly divided into two categories: those designed for straightforward pick-and-place operations with industrial-style grippers and complex robotic hands intended for fine-dexterous in-hand manipulations.
Industrial grippers~\cite{samadikhoshkho2019brief}, such as the Robotiq 2F-85 gripper with two fingers (Robotiq Inc., Saint-Nicolas, Canada), typically feature two opposing fingers to optimize the grasp force, focusing on simple tasks. The Robotiq 3-finger adaptive gripper introduces a third finger for increased adaptability, but remains bulky and limited in versatility~\cite{sadun2016grasping}. 
Researchers in soft robotics have developed various kinds of soft grippers that can be actuated by motors or pressure~\cite{terrile2021comparison}. These grippers are highly adaptable and safe to use, but lack dexterity and applied force. 
The iRobot-Harvard-Yale Hand~\cite{iHY-Hand} and the Eagle Shoal Hand~\cite{Eagle_Shoal} represent efforts to enhance versatility through simple and cost-effective designs. These grippers manage a range of tasks from pinching to power grips but fail to perform complex hand manipulations. 
At the upper end of design complexity, grippers like the Schunk SVH Hand (Schunk SE \& Co. KG, Lauffen am Neckar, Germany) and the Shadow Dexterous Hand (Shadow Robot Company, London, United Kingdom) mimic the physiology of the human hand with multiple actuators and DoF, excelling in the task of in-hand manipulation~\cite{andrychowicz2020learning}. However, their complexity and high cost limit their accessibility.
Attempts to simplify and reduce costs while maintaining functionality have led to designs such as Carnegie Mellon University's LEAP Hand~\cite{shaw2023leap} and Allegro Hand~\cite{WonikRobotics2023}, which use direct actuation for a balance of compliance and control, but still face challenges in dexterity.

The landscape of robotic grippers thus presents a divide between high-cost, sophisticated biomimetic hands and simpler, robust industrial grippers that often compromise on dexterity. A design pipeline addresses this divide with a grammar-based modular system to quickly design task-specific manipulators~\cite{zlokapa2022integrated}. Although modular systems offer flexibility and ease of iteration, they are optimized for narrow use cases, limiting their ability to perform a broad spectrum of complex tasks. Vision-controlled jetting allows for rapid multi-material printing of complex hand designs in one go, and therefore further lower the cost~\cite{buchner2023vision}.

From an application perspective, artificial hands can be categorized into three main sectors: prosthetics and rehabilitation (including assistive robotics), industrial applications, and human-robot interactions. Each sector presents its own unique set of requirements. Although prosthetics and rehabilitation have historically dominated the field of robotic hand applications, industrial and human-robot interaction applications have gained significant attention in recent years.
In response to this growing interest, a design that combines the natural appearance of a human-like hand for assistive purposes with the functional capabilities of industrial grippers could potentially meet the demands of multiple sectors.
To satisfy the different requirements across these sectors, the design should allow for adaptivity without compromising robustness, maintain high efficiency, and remain safe and natural when interacting with humans. Furthermore, achieving simplicity in control is a crucial aspect of the design. These requirements have led to a shift towards simplified designs, such as moving away from fully actuated systems in favor of coupled or underactuated transmission architectures, and pushing novel joint design approaches away from rigid joints towards more flexible or soft joints~\cite{piazza_century_2019}.

\subsection{Contributions}


Our work aims to bridge the gap between biomimetic hands and industrial grippers by offering a versatile design that merges the affordability of industrial models with the functional versatility of more complex systems. This solution achieves proficient performance in both straightforward pick-and-place tasks and intricate in-hand manipulations. Unlike modular approaches, which focus on task-specific customization, it provides a general-purpose solution that maintains dexterity and adaptability across a wide range of tasks through its innovative rotating thumb mechanism.
The contributions of this work are: 
\begin{itemize}
    \item  a new mechanical design of a robotic hand with a rotating thumb that combines the advantages of biomimetic hands and industrial grippers;
    \item  a hollow rolling-contact joint to simplify the joint kinematics; and 
    \item the demonstration and evaluation of the system, using test items from the YCB dataset~\cite{YCBcomplete}.
\end{itemize}

\section{Design}
\label{sub:design}

\begin{figure}[t]
        \centering
        \includegraphics[width=\linewidth]{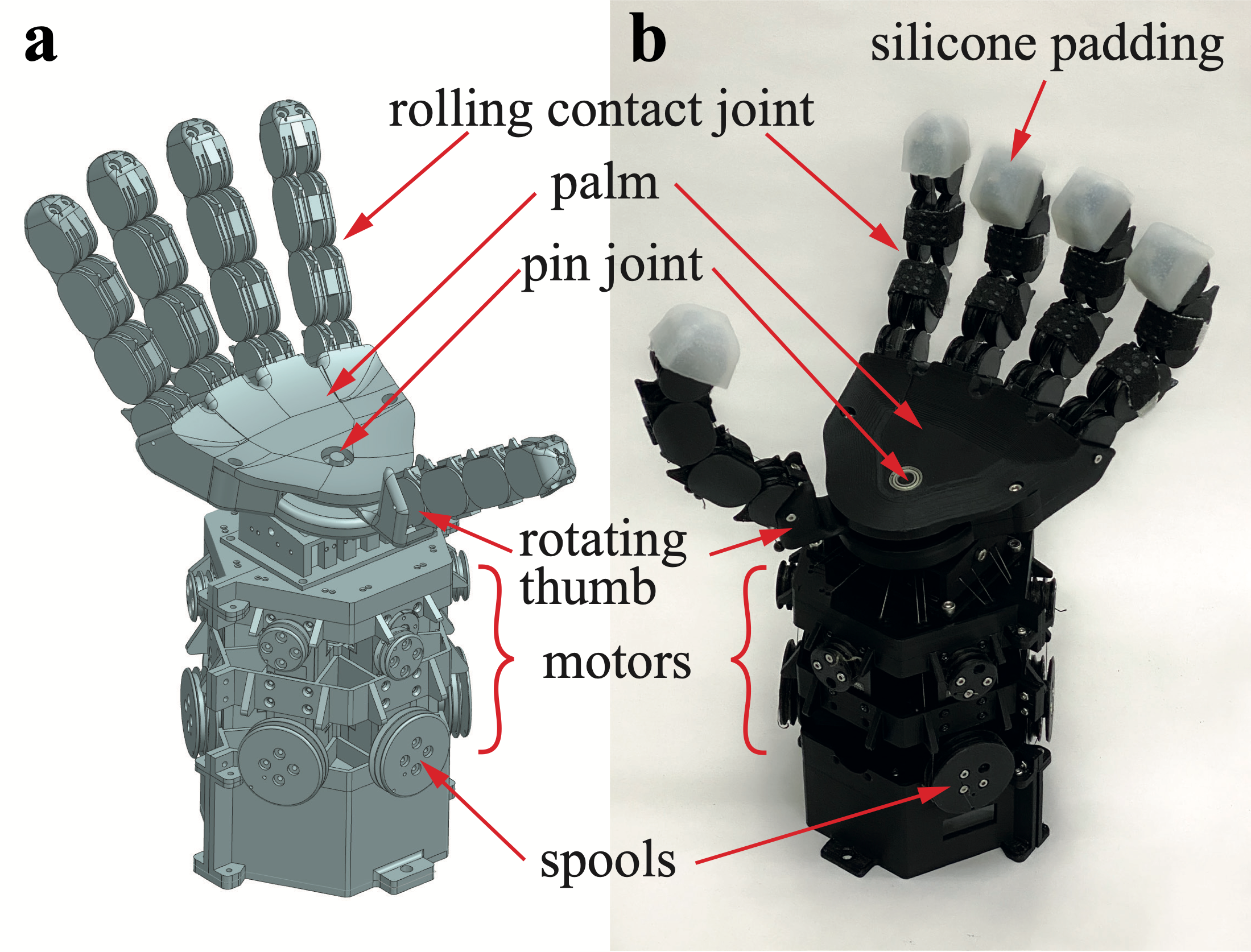}
        \caption{Rotograb's CAD model compared to its real-world instantiation. \textbf{(a)} CAD model of Rotograb with five fingers, including the rotating thumb and the actuation tower below the wrist with motors and spools. \textbf{(b)} Assembled 3D-printed Rotograb with silicon padding for increased contact friction.}
        \label{fig:fullDesign}
\end{figure}

Rotograb aims to fuse the advantages of human hands with industrial grippers. To do so, Rotograb has five fingers to mimic and achieve the dexterity of a human hand. All fingers have the same design to simplify modeling and fabrication. While the index, middle, ring, and pinkie are attached to the palm, the thumb can rotate around the palm to switch between different configurations: right hand, left hand, or middle (opposing the other fingers). It allows the hand to switch between a 'dexterous' and a 'powerful' grasp. The hand has 11 DoF, distributed as follows: two per finger and one for the rotating plate. The hand is mounted with an angle of $10^{\circ}$ with respect to the horizontal wrist plane where 11 servo actuators are located. \Cref{fig:fullDesign} shows the details of Rotograb as well as its CAD model.

\subsection{Fingers}
\label{sub:finger}
\begin{figure}[t]
        \centering
        \includegraphics[width=\linewidth]{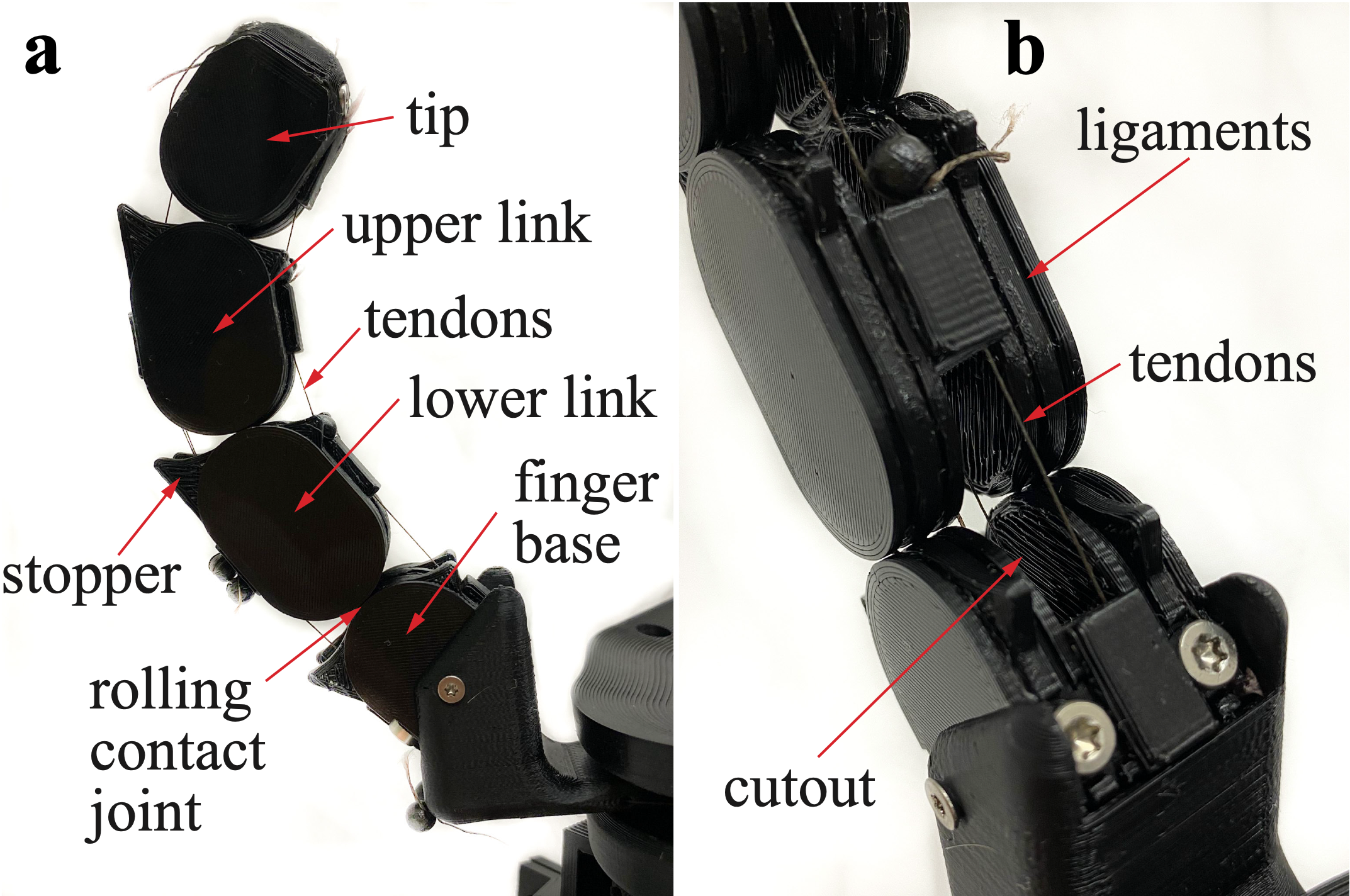}
        \caption{Uniform design and functionality of the hand's robotic finger with a base, three linked segments, rolling-contact joints, and a joint cut-out. \textbf{(a)} Side view of a finger, specifically the thumb. All fingers have the same design. The finger consists of a finger base and three links (lower link, upper link, and tip). The links are connected by rolling-contact joints and actuated through tendons. Mechanical stoppers limit the movement. \textbf{(b)} Close-up of the rolling contact joint with tendons and ligaments. The cut-out in the center of the rolling-contact joint simplifies the finger kinematics.}
        \label{fig:closeup_thumb}
\end{figure}

All fingers follow the same design. Each finger is made up of four links: the base, the lower and upper link, and the tip (\Cref{fig:closeup_thumb}). The links are connected by three rolling contact joints. In particular, they are held together by four separate ligaments attached in alternating directions (top and bottom sides of the finger) to have symmetric flexibility along the lateral plane. Rolling contact joints minimize friction and provide some lateral compliance to the fingers. In addition, when they are subjected to excessive stress, they pop instead of breaking. The aforementioned benefits of the rolling contact joint come at the cost of a lower control accuracy.

\Cref{fig:Fingers_cuts} shows the tendon routing in details. From a 2D lateral perspective, each part of the finger is made from the combination of a rectangle and two half circles with the same radius. The movement generated by the rolling contact joint results from two simultaneous rotations of equal magnitude about distinct axes. To better understand this concept, one can consider the rotation of the lower link over the finger base in \Cref{fig:Fingers_kinematics}. The 'virtual' rotation refers to the rotation of the link around the axis passing through the point $O_1$, while the 'real' rotation occurs around the axis passing through $O_1'$. It is essential to note that the two contact surfaces are obtained from cylinders with identical radii. In this 2D representation, they appear as semicircumferences centered at $O_1$ and $O_1'$. The equality of the radii explains why the rotations around $O_1$ and $O_1'$ are of equal magnitude, as they are coupled by the ligaments and share the same amplitude.

The base of the finger is attached to the palm (or to the rotating plate) at an angle of $45^{\circ}$. Joint~1 allows the finger to rotate backwards up to $-45^{\circ}$ (flat position), while joints~2 and~3 can only flex forward. The maximum inward flex angle is $90^{\circ}$ for all joints. Mechanical stoppers enforce the range of motion.

\begin{figure}[t]
        \centering
        \includegraphics[width=\linewidth]{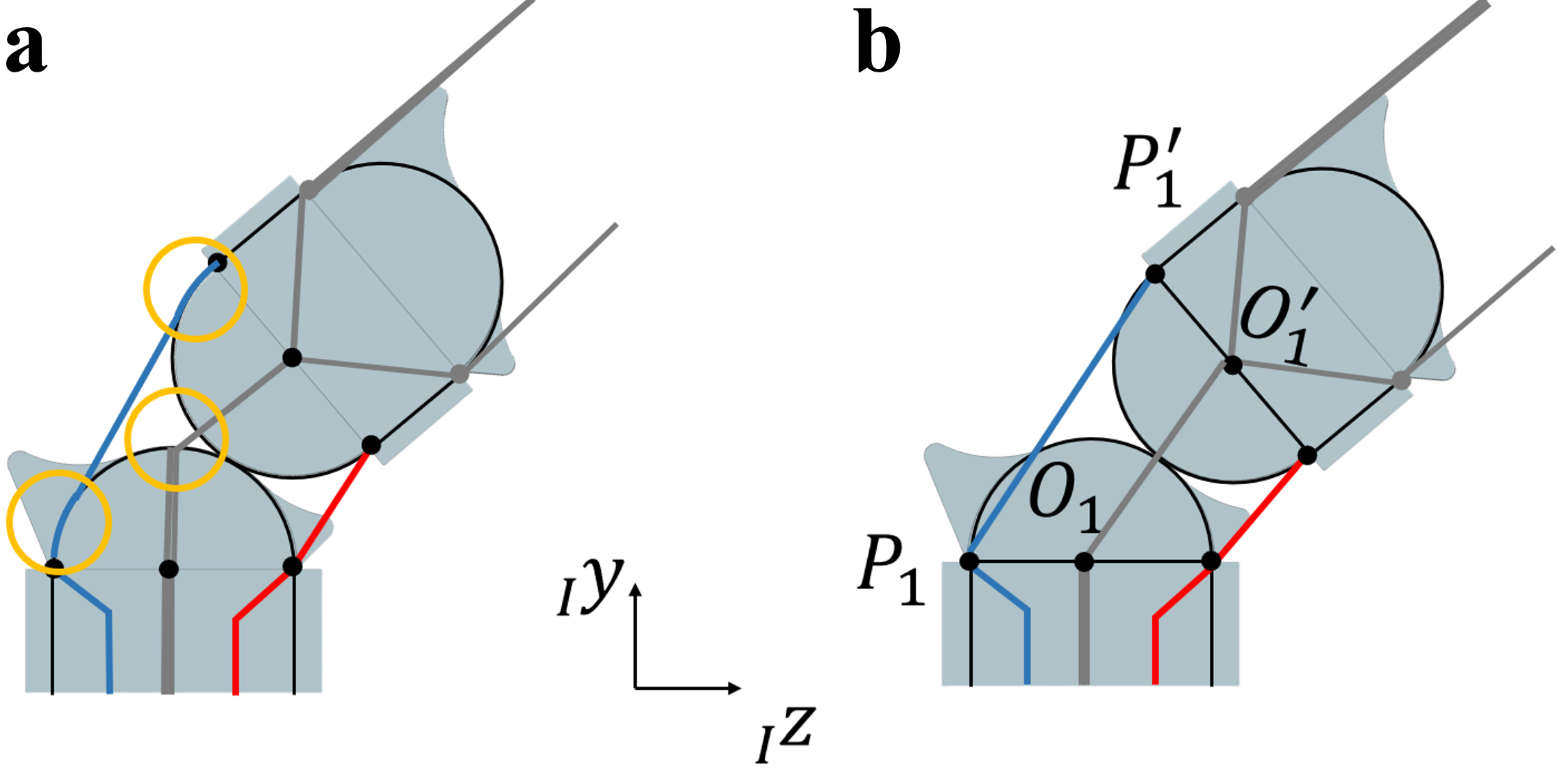}
        \caption{Comparison of a normal rolling contact joint \textbf{(a)} and our rolling contact joint with the previously named cut-out \textbf{(b)}. In the normal rolling contact joint, the tendon wraps around the joint while actuating the finger, extending the tendon length. In the adapted rolling contact joint, these critical non-linearities (yellow circles) are bypassed.}
        \label{fig:finger_cutout}
\end{figure}

Tendon length calculations of the tendon routing are complex for rolling contact joints because the contact point from which the tendon applies a force to the link changes with the angle. To simplify the problem, we cut out the internal part of each link. This results in simplified control and kinematics with a minimal impact on robustness, as discussed in \Cref{section:kinematics}. To elucidate the design of the cutout, we focus on joint~1 and the circles centered at $O_1$ and $O_1'$ as depicted in \Cref{fig:finger_cutout}. The cutout for this joint is clearly shown in \Cref{fig:closeup_thumb}. The tendons associated with joint 1 are external and act directly from the extremity of the circles' diameter. The tendons for the upper joint are routed through $O_1$ and $O_1'$, remaining unaffected by any changes in the angle of joint 1. In both cases, the cutout is essential as it prevents the tendons from following the curvature of the circle.

The tips of all fingers are encased in silicone. This wrapping increases the friction coefficient to grasp objects more easily and provides a higher degree of softness for dealing with delicate objects.

\subsection{Rotating Thumb}
\label{sub:rot_thumb}

The thumb is designed in the same way as the other fingers and is fixed on a rotating plate. The plate is attached to the center of the palm through a pin joint with a bearing to minimize friction. The resting position is shown in \Cref{fig:fullDesign}.a). In this configuration, it acts in the opposite direction to the other fingers.

The plate is actuated using two tendons that pull on the right and left side, respectively, and can move from $-65^{\circ}$ to $+65^{\circ}$. The tendons that act on the thumb itself are routed through the interior of the plate. To decouple the flexion and extension of the thumb from the rotation of the plate, the thumb tendons are routed as closely as possible to the axis of rotation of the plate. This ensures that the angle of the plate does not affect the tendon lengths of the thumb.

\subsection{Wrist}
\label{sub:wrist}
The hand is mounted at an angle of $10^{\circ}$ with respect to the plane orthogonal to the axis of the wrist. This choice allows the hand to grasp objects from above and operate in limited space, similar to the eagle's talon. The orientation of the hand minimizes the torque applied to the wrist, since most of the gravitational force is applied along the axis of the wrist. 

\subsection{Actuation and Tendon Routing}
\label{sub:routing}
\begin{figure}[t]
        \centering
        \includegraphics[width=\linewidth]{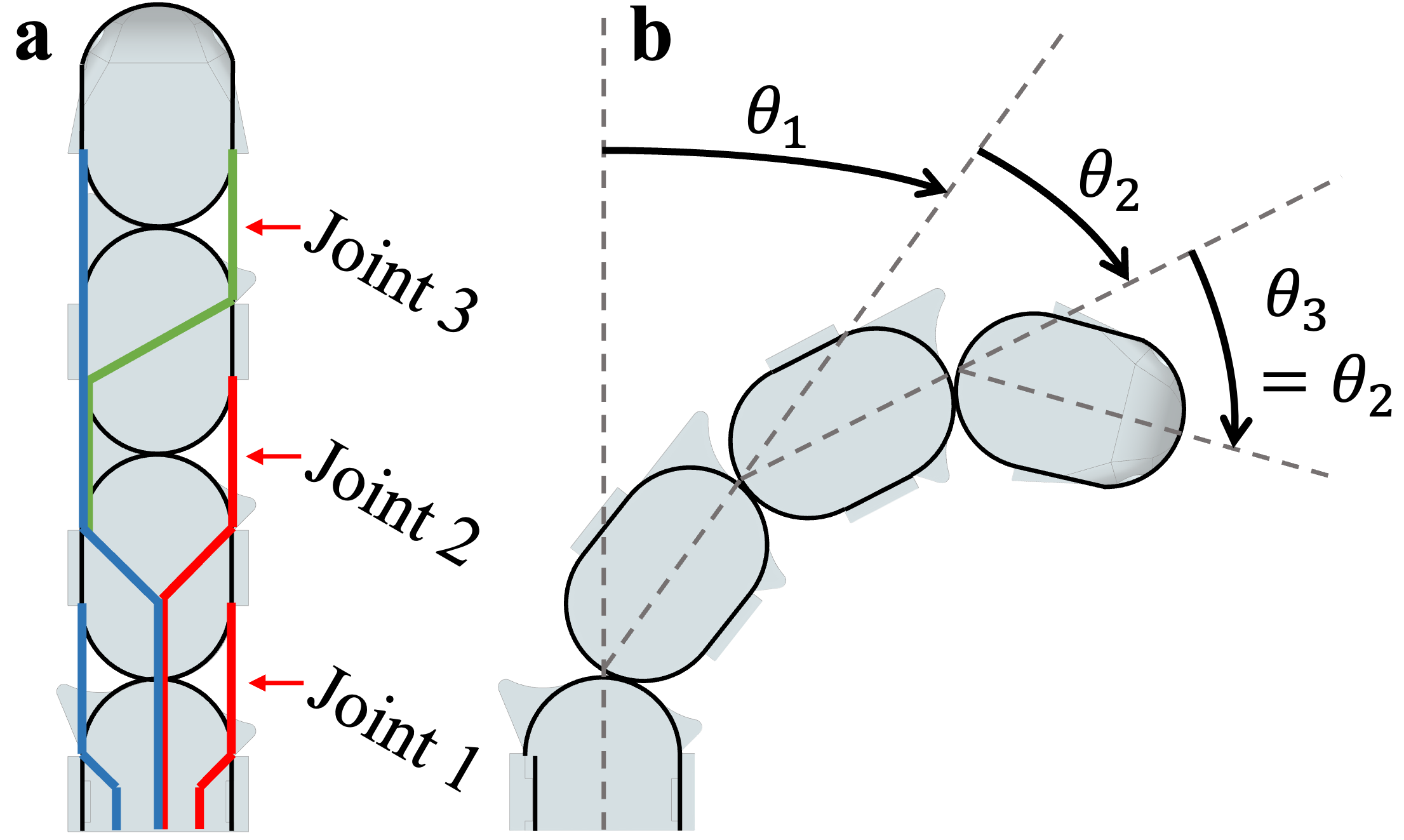}
        \caption{Detailed representation of a finger's internal flexor and extensor tendon routings, illustrating the coupled movement of two joints to demonstrate the complex mechanics of finger articulation. \textbf{(a)} CAD side view of the tendon routing inside the finger with three joints. Flexor tendons are drawn in red, extensor tendons in blue, while the green tendon couples the flexion of joints 2 and 3. \textbf{(b)} Illustration of the joint angles of a finger, with joints 2 and 3 being coupled.}
        \label{fig:Fingers_cuts}
\end{figure}

Each tendon is attached to a servomotor through a spool. The tendons are routed inside the wrist to the fingers. Inside the finger, the routing follows the routing of human tendons with two flexor and two extensor tendons~(\Cref{fig:Fingers_cuts}.a). The contact point of the tendons, from which they pull the links, is always chosen to be at the extremity of the diameter of the circles of the links. 

The tendons are routed through the cutout of the links. A key benefit of this is that the distance between the centers of rotation of two adjacent links is constant (e.g., segment $O_{1}O_{1}'$ in \Cref{fig:Fingers_kinematics}.a). To decouple the actuation of the lower and upper parts of the finger, we route the tendons for joints 2 and 3 through the centers of rotation of joints 1 and 2, respectively. In this way, the length of the tendon of the upper joints will not be affected by the actuation angle of the lower joints (see \Cref{section:kinematics} for more details).

Each finger is actuated by two motors. One controls joint~1, while the other controls the two coupled upper joints~2~and~3. Each motor is attached to two spools, one for the flexor and one for the extensor tendon. Due to the different amounts of tendon pulled in the flexor and the extensor for the upper coupled joints, the spools are designed with varying radii that maintain a constant ratio. The extensors for joint 2 and joint 3 are combined in one tendon. This combination makes the single extensor increase/decrease twice as much as the corresponding flexor of joint 2. Choosing to double the radius of the spool for the extensor rather than for the flexor simplifies kinematic calculations compared to those introduced in \Cref{section:kinematics}.

\subsection{Materials and  cost}
The total material costs of Rotograb are approximately 1100\,€. The largest part is for the 11 servo actuators. The remaining expenses cover the control and power units, as well as the tendon material, silicon, ball bearing, screws, and bolts.
We used 3D printed material for the core structure (palm, wrist, and fingers), which allowed us to implement changes in the design in a quick and inexpensive manner and, therefore, is ideal for designing a proof-of-concept as the Rotograb hand.
Furthermore, the cost of the camera, which we use for control, starts at 316\,€.

The detailed costs of the key hardware components are listed in \Cref{tab:components_table}.

\begin{table}[ht]
\caption{List of components used in the project.}
\centering
\renewcommand{\arraystretch}{1.5}
\begin{tabularx}{0.48\textwidth} { 
   >{\centering\arraybackslash}p{1.8cm} 
   >{\centering\arraybackslash}p{1.2cm}
   >{\centering\arraybackslash}X
   >{\centering\arraybackslash}p{0.9cm}  }
 \toprule
 \textbf{Component} & \textbf{Supplier} & \textbf{Model} & \textbf{Cost}\\
 \midrule
 \small Servo Motors  & \small Robotis  & \small XC330-T288-T & \small 896\,€\\
\small Control Unit & \small Robotis & \small U2D2 & \small 81\,€\\
\small Power Hub & \small Robotis & \small U2D2 Power Hub & \small 45\,€\\
\small Power Supply & \small Robotis & \small SMPS 12V 5A PS-10 & \small 50\,€\\
\small Camera &  & \small OAK-D Pro AF & \small 316\,€\\
\bottomrule
\end{tabularx}
\label{tab:components_table}
\end{table}

The total height of the wrist and hand is 27.5\,cm. The length of the fingers and the thumb is 9.6\,cm, and the width of the palm is 9.4\,cm. The total length of the hand (measured from palm to fingertip) results in 17.9\,cm. These dimensions are comparable with those of average male hands. Therefore, Rotograb can be used to mimic human interactions with objects.

\section{Kinematics}
\label{section:kinematics}
The hand is fully actuated by tendons through eleven servomotors placed on the wrist. We made three design choices to simplify the kinematic calculations. First, the inside of the links is removed. Therefore, the flexor and extensor are routed through the center of the rolling contact joint. This decouples the bending of different joints. A bending in the lower joint (e.g., joint 1) does not affect the tendon actuation of the upper joint (e.g., joint 2).  Therefore, we can calculate the deflection only at the actuated joint. Any nonlinearities of bending tendons are bypassed. Second, the tendons of the thumb are routed through the center of rotation of the plate. In this way, the rotation of the plate does not affect the tendon lengths of the flexors or extensors of the thumb. Third, the choice of spool radii takes into account the increase and decrease of differences in tendon length.

\subsection{Fingers}
\label{sec:Finger_joint_kin}
Each individual finger has two actuated DoFs with the same geometry and tendon routing. We use a simplified 2D problem to compute the finger kinematics, which is the same for all fingers. 

To actuate the robotic hand, the desired joint angle $\theta$ must be translated into a change in the length of the tendon $\Delta l$. Divided by the radius of the spool, this results in the motor rotation needed to achieve the desired joint angle. Since the length of the tendon, which is routed through the palm and the interior cutout of the links, stays constant, the only varying lengths are found between consecutive links, denoted as $\Vec{l}_{P_{i}P_{i}'}$.

\begin{figure}[t]
        \centering
        \includegraphics[width=\linewidth]{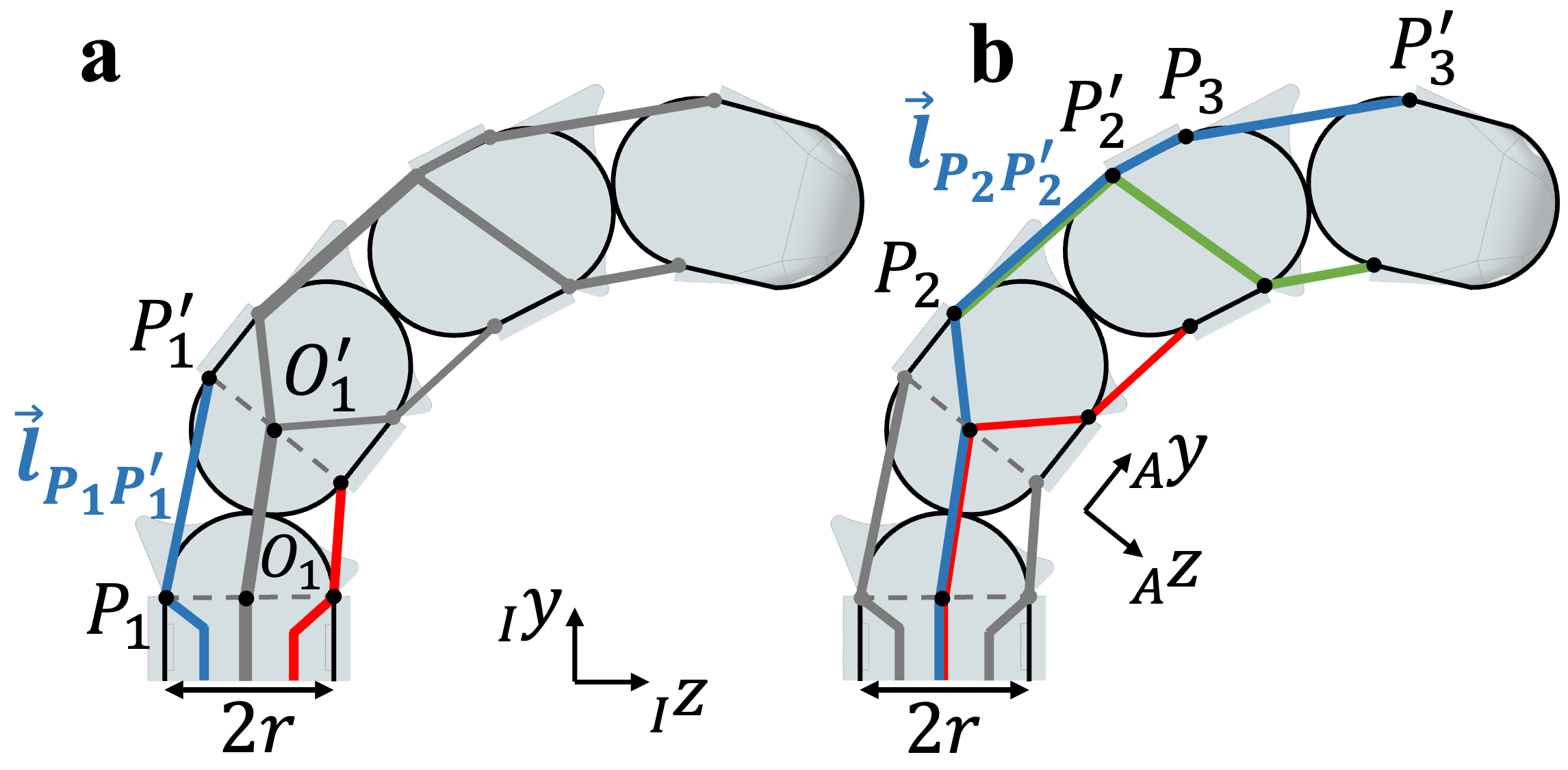}
        \caption{2D model of the finger kinematics shown separately for the lower and upper sections. \textbf{(a)} Joint 1 modeled with flexor (red) and extensor (blue) tendons. \textbf{(b)} Joints 2 and 3 with combined flexor (red) and extensor (blue) tendons and the coupled flexor tendon (green).}
        \label{fig:Fingers_kinematics}       
\end{figure}

\begin{figure}[ttt]
    \centering
    \includegraphics[width=0.7\linewidth]{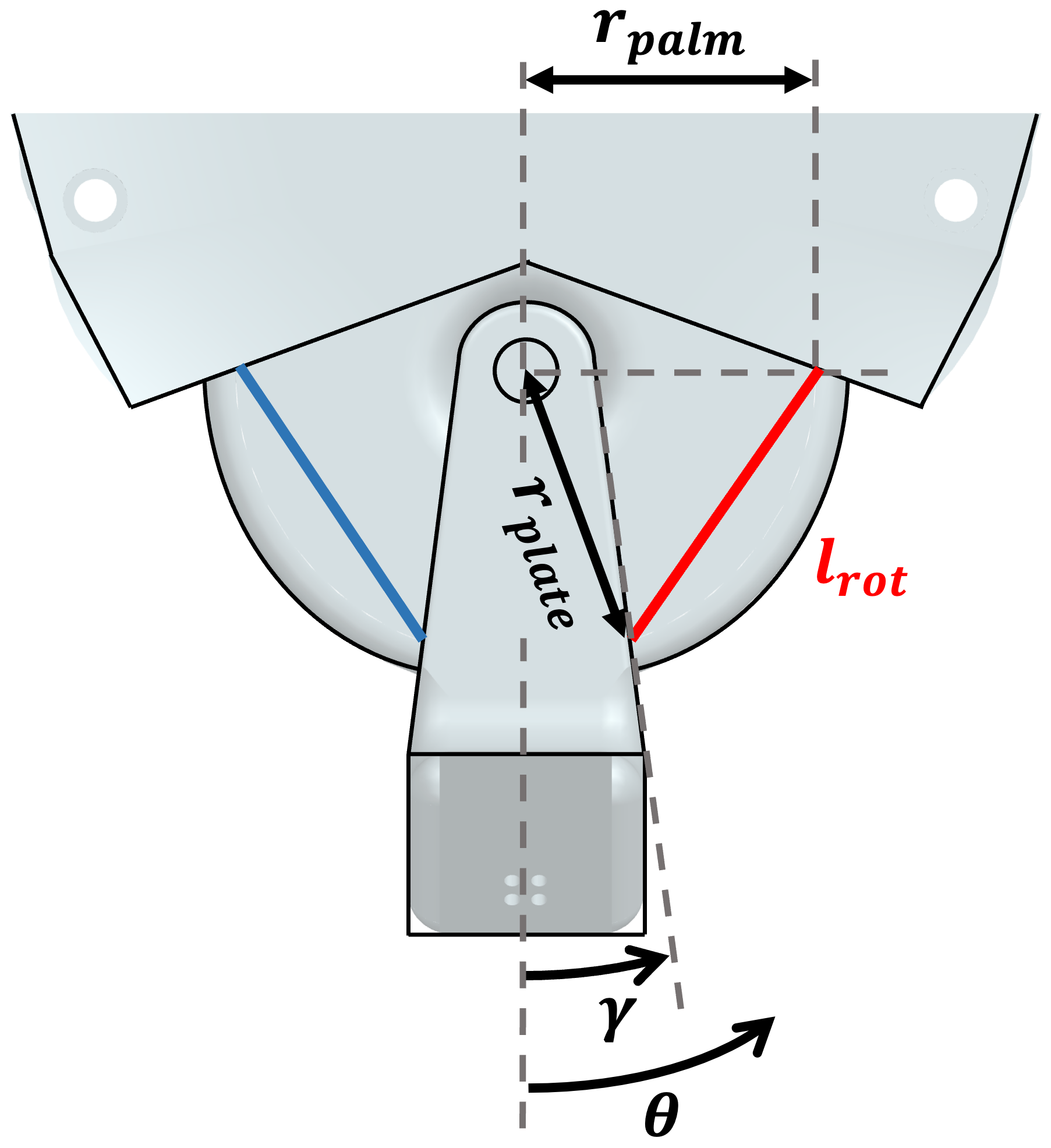}
    \caption{2D model of the rotating thumb.  $\theta$ indicates the state of the thumb's rotation, with $\theta=0$ indicating the middle position of the thumb.}
    \label{fig:Kinematics_thumb}
\end{figure}

\begin{figure*}[t]
    \centering
    \includegraphics[width=\linewidth]{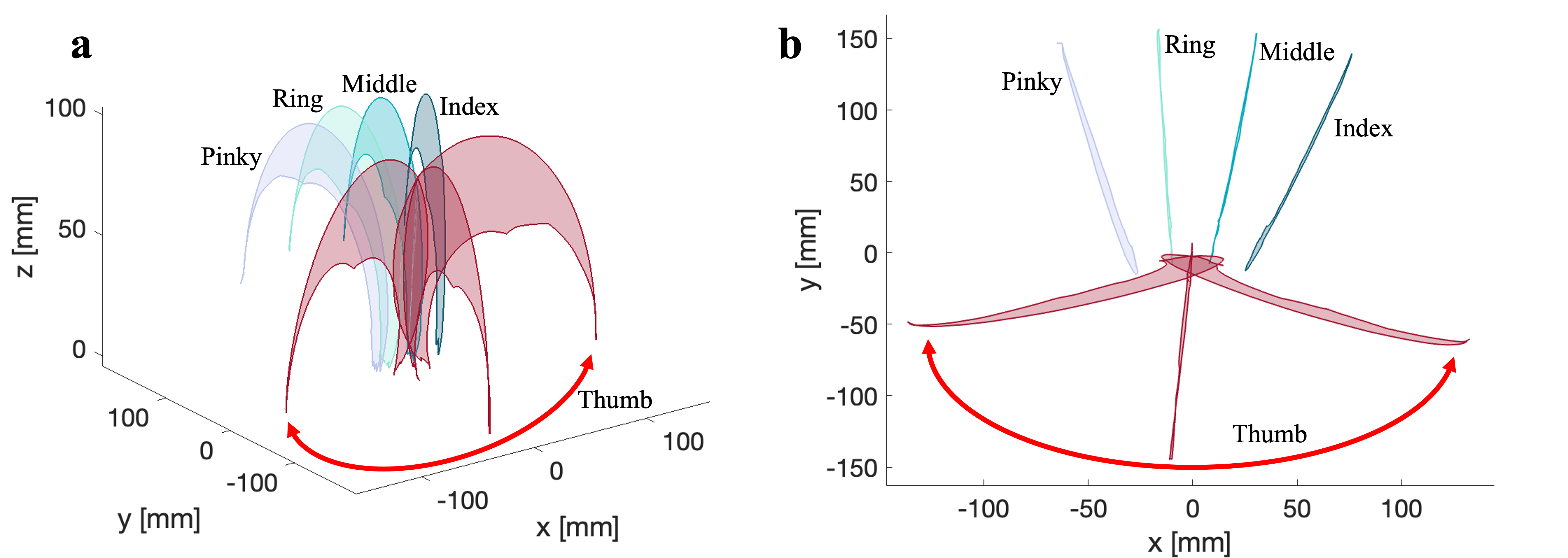}
    \caption{Workspace analysis of Rotograb's five fingertips. The plots indicate the outer boundary of the fingertip motions in \textbf{(a)} 3D and \textbf{(b)} projected onto the x-y-plane (orthogonal to the center line of Rotograb's actuation tower). The fingers are indicated in the plot, and the thumb is shown in the left, middle, and right positions. The fingers are named from the perspective of the right hand.}
    \label{fig:workspace}
\end{figure*}

The equations for the kinematics of the finger are introduced by first looking at joint 1 using the model depicted in \Cref{fig:Fingers_kinematics} (a). The points $O_{1}$ and $O_{1}'$ are the centers of the virtual and real rotations, respectively. The distance between these points remains the same as a result of the design features mentioned at the beginning of this section. The points $P_{1}$ and $P_{1}'$ are the exit points and entry points of the extension tendon, respectively. For calculations, we only consider the length of these two points.

The calculation of the vector of tendon length $_{I}l_{P_{1}P_{1}'}$ can be simplified to the following equations:

\begin{equation}
\begin{split}
        _{I}l_{P_{1}P_{1}'}(\theta_{1}) &=
    \begin{bmatrix}
        _{I}l_{y}(\theta_{1}) \\
        _{I}l_{z}(\theta_{1}) \\ 
    \end{bmatrix} \\
    &=
    \begin{bmatrix}
        r\cdot\sin\theta_{1} + 2r\cdot\cos\frac{\theta_{1}}{2} \\
        r\cdot(1-\cos\theta_{1}) + 2r \cdot \sin\frac{\theta_{1}}{2} \\
    \end{bmatrix}
\end{split}
\end{equation}

The final change in tendon length $\Delta l(\theta_{1})$ is calculated by
\begin{equation}
    \Delta l(\theta_{1}) =  \lVert _{I}l_{P_{1}P_{1}'}(\theta_{1}) \rVert - \lVert _{I}l_{P_{1}P_{1}'}(\theta_{1,init}) \rVert
\end{equation}
$\theta_{1}$ is the desired joint angle chosen for joint~1, and $\theta_{1,init}$ is the calibration angle. In the case of joint~1, it holds that $\theta_{1,init} = -45^{\circ}$. To move the finger, the change in tendon length is added to the extensor length and subtracted from the flexor length.

Due to design choices, the calculations for joint~2 are done analogously to those for joint~1. As seen in \Cref{fig:Fingers_kinematics}.b, for joint~2, we use the reference frame $A$ and apply the same calculations to obtain the vector of the tendon length $_{A}l_{P_{2}P_{2}'}$ and the change in tendon length $\Delta l(\theta_{2})$ depending on the desired joint angle $\theta_{2}$ and the calibration angle of joint~2, $\theta_{2,cal}=0^{\circ}$
The flexion or extension of joint~3 happens simultaneously with joint~2 since the coupling connects movements directly.

\subsection{Rotating Thumb}

Calculations of the tendon length of the rotational thumb can also be simplified to a 2D problem, as seen in \Cref{fig:Kinematics_thumb}. Knowing the dimensions $r_{palm}$, $r_{plate}$, and $\gamma$, the tendon length $l_{rot}$ is computed using the cosine theorem:

\begin{equation}
    l_{rot}(\theta) = \\ \sqrt{r_{palm}^{2}+r_{plate}^{2}-2 r_{palm}r_{plate}\cos\left(\frac{\pi}{2}-(\theta+\gamma)\right)}
\end{equation}

The change in tendon length is then computed by:
\begin{equation}
    \Delta l_{rot}(\theta) = l_{rot}(\theta)-l_{rot}(\theta_{0})
\end{equation}

For calibration purposes, we choose $\theta_0 = 0$. Due to the symmetry of the design, the length of the left tendon is increasing by the same amount as the length of the right tendon is decreasing.

The tendon routing of the thumb goes through the axis of rotation of the plate, designed to decouple the hinge joint in the palm and the flexion/extension of the thumb. Thus, the calculation for the thumb's tendons can be performed analogously to the one for the other fingers (see \Cref{sec:Finger_joint_kin}).

\section{Control}
\label{section:teleoperation}

\subsection{Teleoperation}
Rotograb can be operated through teleoperation. The user's hand movements serve as the input, which is then mapped to the gripper.
We used an OAK-D Pro (Luxonis, Littleton, CO, United States) depth camera to capture the movements of a human user. The camera tracks the user's hand, and then the landmarks are extracted using the Google Mediapipe hand tracking model~\cite{lugaresi2019mediapipe}.
The input angles of the human hand are mapped for each actuated joint to the robotic gripper. For the index, middle, ring, and pinky fingers, the mapping is done directly. 
For the thumb, it detects whether the human is showing the left or the right and consequently switches the hand's mode.
All angles are scaled and tuned to achieve smooth operation of the robotic hand over its full workspace, especially for the workspace of the rotating thumb.

\begin{figure*}[t]
    \centering
    \includegraphics[width=\linewidth]{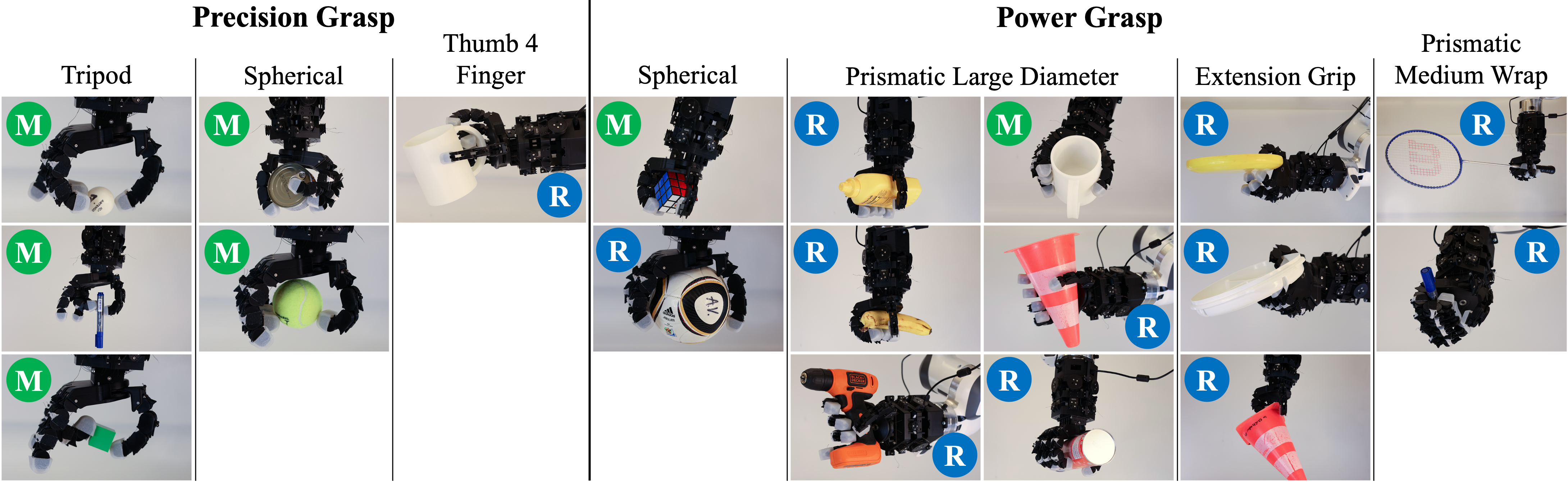}
    \caption{Rotograb grasping different objects from the YCB dataset~\cite{YCBcomplete} with precision or power grasps. The grasps are categorized by the grasp taxonomy introduced by Mark R. Cutkosky et al.~\cite{cutkosky1989grasp}. We indicate the position of the thumb by ``L'', ``M'', or ``R'', meaning left, middle, or right position, respectively.}
    \label{fig:objects}
\end{figure*}

\begin{figure}[t]
        \centering
        \includegraphics[width=\linewidth]{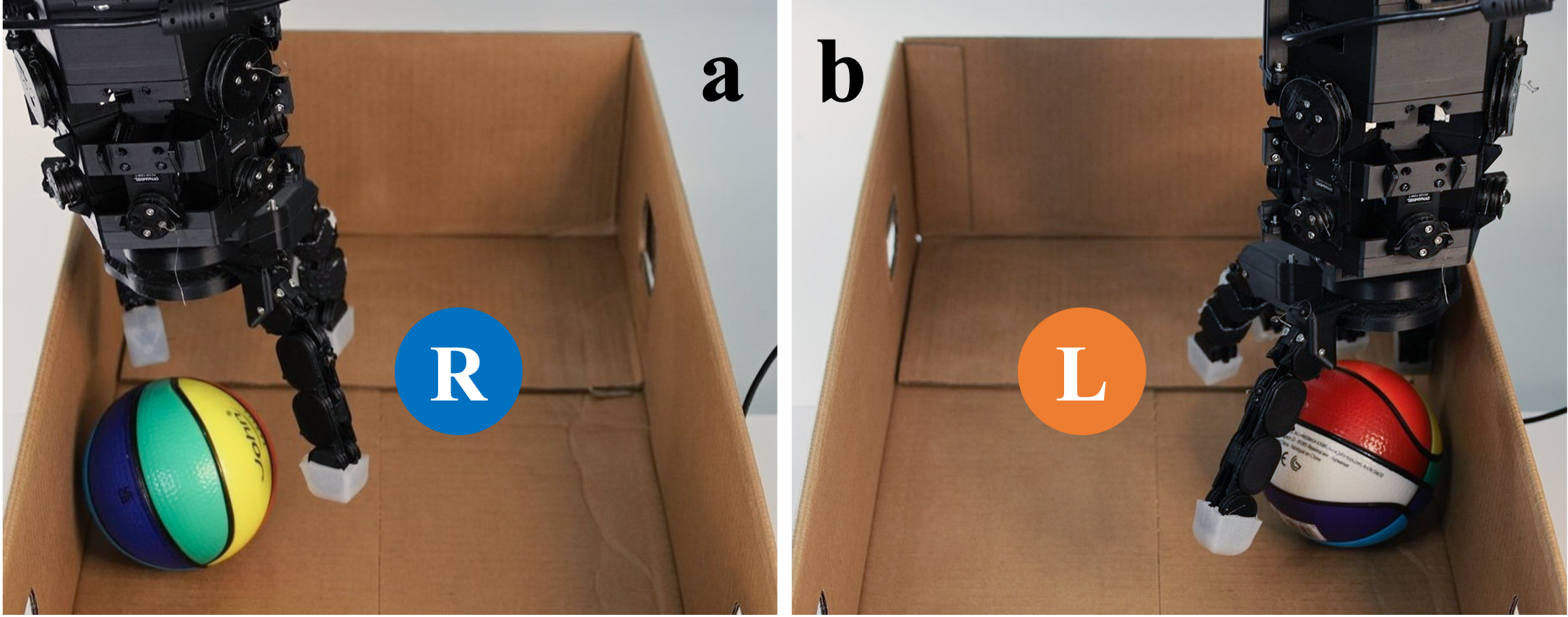}
        \caption{Showcase of the ambidexterity of Rotograb. The hand grasps a ball that is constrained by a closed wall. \textbf{(a)} The wall on the left side of the ball. Rotograb can grasp a ball with the thumb on the right side. \textbf{(b)} The wall is right of the ball. Rotograb can grasp a ball with the thumb on the left side.
        }
        \label{fig:ambidexterity}
\end{figure}

\begin{figure}[t]
        \centering
        \includegraphics[width=\linewidth]{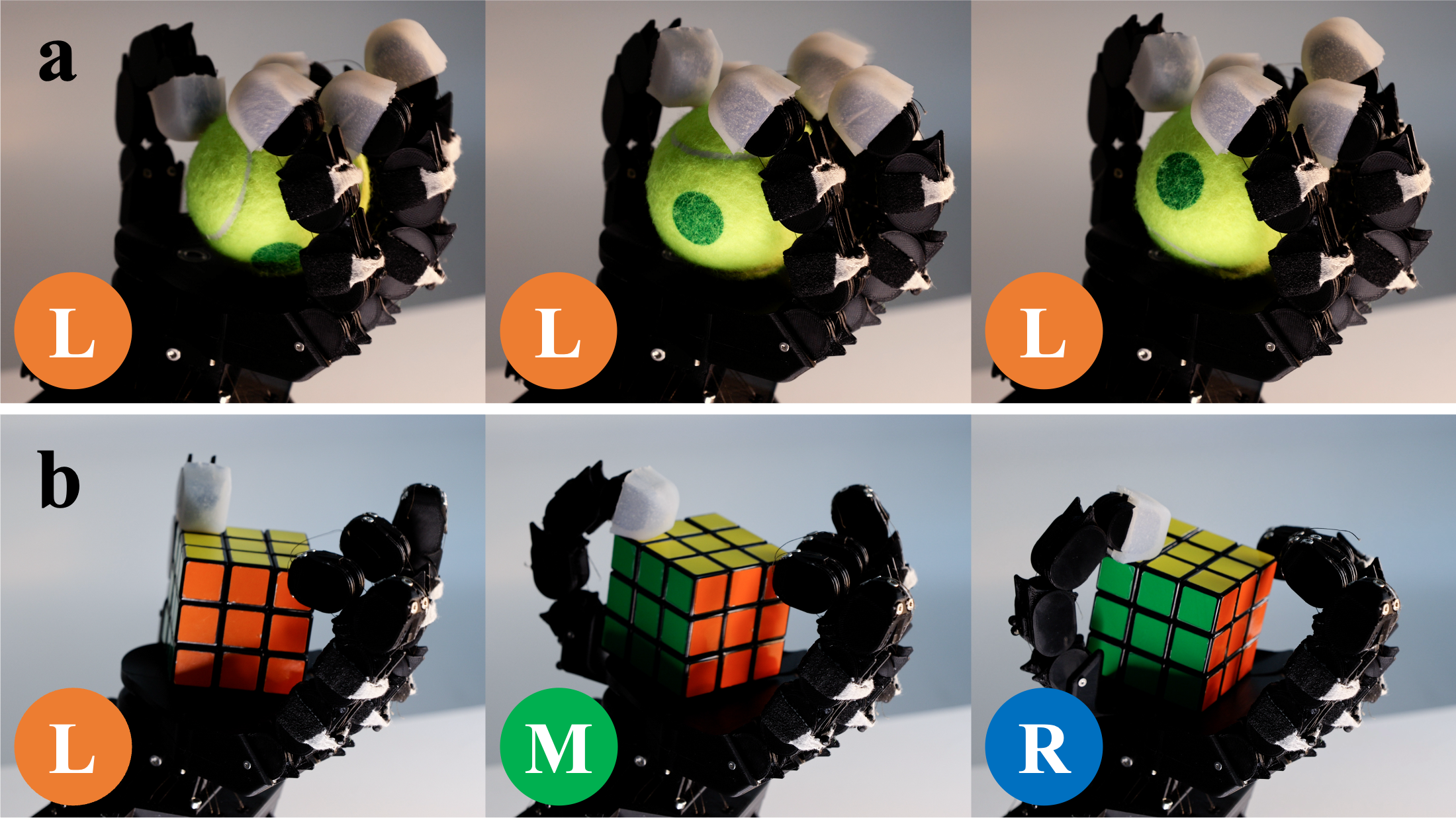}
        \caption{Showcase of the in-hand manipulation capabilities of Rotograb, based on Reinforcement Learning. \textbf{(a)} Rotograb rolls a ball with the thumb on the left side and the other four fingers doing most of the ball rolling. \textbf{(b)} Rotograb re-orients a Rubik's cube using its rotating thumb.}
        \label{fig:InHandRotation_sequence}
\end{figure}

\subsection{Reinforcement Learning for In-Hand Object Rotation}

We implemented a reinforcement learning pipeline to rotate objects in the hand autonomously.
We follow a similar pipeline to Toshimitsu et al.~\cite{toshimitsu2023getting}, using the proximal policy optimization (PPO) algorithm~\cite{schulman2017proximal} implemented by \textit{rl\_games}~\cite{rl-games2021}. We trained to roll a ball on the palm in two opposite directions and also tested the hand with other objects.
With the GPU-based IsaacGym simulator~\cite{Makoviychuk2021-ll}, we simulated thousands of robots in parallel. All dexterous manipulation policies were learned in around 40 minutes on a single NVIDIA GeForce RTX 3060 GPU. Each training is 2000 epochs long and involves 4096 simultaneous environments.

The action space used in Isaac Gym consists of the joint angular velocities, while the observations available in the simulation are the true joint angles and angular velocities, as well as the position, orientation, velocity, and angular velocity of the manipulated object. The specific reward for ball rotation is based on the object's rotational velocity, which returns the maximum reward when the absolute rotational velocity around the x axis is between 1 and 3\,rad/s and linearly decreases outside of this region. The sign of $\omega x$ in the reward formula can be flipped to reverse the desired direction of rotation.

To compensate for the inaccuracy of the physics engine and make the policy more robust to overcome the sim2real gap, we applied domain randomization to the physics properties. Gaussian or uniform noise was added to the observations, actions, damping and stiffness of the tendons and joints, joint range of motion, mass and friction of the robot and object, and object scale.

Since sensor feedback was not implemented on the physical system, the policy was recorded during the simulation as a sequence of joint angles and executed in a feedforward manner on the physical system.
Possible improvements to this approach would include using a camera to measure ball position and incorporating velocity feedback into policy execution. Further enhancement could also involve estimating the joint position and velocity using a Kalman filter, IMUs, and angular deflection sensors.

\section{Experimental Evaluation}

\subsection{Workspace}

To quantify how Rotograb can reach fingertip positions in its workspace, we analyzed the fingertip motions in a motion capture system (Qualisys). \Cref{fig:workspace} shows the workspace of the different fingers in the 3D space and projected to the x-y plane.
The rotating thumb captures a larger workspace than the other fingers. This shows that our design philosophy can indeed capture different states, from industrial grippers to dexterous hands with ambidexterity. 
We also see that each finger captures a similar workspace. 
For future improvements, we will add adduction and abduction to all fingers to close the workspace gap between the fingers.

\subsection{Object Grasping}
Rotograb was tested with a set of objects chosen from the YCB benchmark dataset~\cite{YCBcomplete} to assess the grasping capabilities. The objects used for the experiment are depicted in \Cref{fig:objects}. We find that Rotograb can grasp small, delicate objects like a ping pong ball or a pen, as well as large objects such as a handball or pylons. Depending on the object, Rotograb switches between a human-hand state (thumb on the left or right side) or an industrial gripper state (thumb in the middle position).
All grasping tasks reported in this paper are performed through teleoperation.

\subsection{Ambidexterity}
To evaluate Rotograb's ambidexterity capabilities, we placed a ball in two different positions in a box: next to the left wall and next to the right wall. Therefore, the box walls restrict the grasping procedure, depending on the ball's position. The success of grasping the ball in each position is surveyed for the left and right thumb positions.

As shown in \Cref{fig:ambidexterity}, the two hand modes succeed for different ball positions. It highlights that the design of Rotograb allows for grabbing objects in different situations. The left-hand mode is particularly useful if the grasping is restricted by a wall to the left of Rotograb, while the right-hand mode is usable in configurations where Rotograb is blocked by a wall to the right. This shows that the Rotograb can access objects in confined spaces without changing the rotation or orientation of the robotic arm on which it is mounted.

 \addtolength{\textheight}{-2.5cm}   
                                  
\subsection{In-Hand Manipulation}

In-hand reorientation is a crucial capability for a robot manipulator to perform tasks in real-world scenarios. It requires precise control and sensitivity to design choices. To demonstrate this concept, we present two experiments that underscore the benefits of our rotating thumb design.
As illustrated in \Cref{fig:InHandRotation_sequence}, the experiments involve rotating a tennis ball around the x-axis and a Rubik's cube around the z-axis. Due to its simpler and more symmetric shape, the tennis ball allows for an autonomously learned policy via reinforcement learning in Isaac Gym. This policy involves the thumb alternating between a lateral and a central configuration. Applied to the physical system, the tennis ball rotates at an approximate speed of 5.45 rounds per minute.

For the Rubik's Cube, the policy was hard-coded to emphasize the pivotal role of the thumb. The rotation around the palm proves to be particularly useful for rotating objects around the palm's axis when the abduction movement of other fingers is not feasible. The thumb rotates by pushing the cube's edges, while the other fingers provide countertorque, enabling the rotation. On actual hardware, the cube rotates at an approximate speed of 5.62 rounds per minute.

\section{Conclusion}
Our work presents a new type of robotic hand, called Rotograb, which combines the advantages of humanoid hands with the strength of an industrial gripper. The movable thumb significantly enhances grip strength, particularly for large objects, while maintaining Rotograb's dexterity for small and delicate objects.
We introduce a new type of rolling-contact joint with a cutout in the center to simplify finger kinematics.

Our workspace analysis underlines the potential of the rotating thumb. However, adding adduction and abduction to the fingers will improve the workspace in the future. 
We showcase its capabilities by manipulating a large set of objects, from small pens and ping-pong balls to mugs and drills.
The movable thumb also allows for ambidexterity when performing tasks with a left or right hand. 
We integrate teleoperation control and a reinforcement learning approach. With the latter, Rotograb could manipulate different objects in-hand, such as a tennis ball and a Rubik's cube.

In the next step, we want to increase Rotograb's robustness and make slenderer fingers with adduction and abduction. We will also further enhance control and teleoperation, focusing on the way we map the user input to Rotograb. The differences in shape and size between the human hand and the robotic hand make mapping particularly challenging. 
Furthermore, we want to undertake more quantitative manipulation experiments to analyze and improve Rotograb's capabilities.


\section*{ACKNOWLEDGMENT}
We are grateful for the funding through the SNSF Project Grant \#200021\textunderscore\,215489. The authors thank the team behind the \href{http://www.rwr.ethz.ch}{Real-World Robotics} course at ETH~Zurich, which greatly enabled this project.


\printbibliography

@article{buchner2023vision,
  title={Vision-controlled jetting for composite systems and robots},
  author={Buchner, Thomas JK and Rogler, Simon and Weirich, Stefan and Armati, Yannick and Cangan, Barnabas Gavin and Ramos, Javier and Twiddy, Scott T and Marini, Davide M and Weber, Aaron and Chen, Desai and others},
  journal={Nature},
  volume={623},
  number={7987},
  pages={522--530},
  year={2023},
  publisher={Nature Publishing Group UK London}
}

@misc{WonikRobotics2023,
    title = {Allegro Hand: Highly Adaptive Robotic Hand for R\&D},
    author = {{Wonik Robotics}},
    year = {2023},
    url = {https://www.wonikrobotics.com/research-robot-hand}
}

@misc{rl-games2021,
title = {rl-games: A High-performance Framework for Reinforcement Learning},
author = {Makoviychuk, Denys and Makoviychuk, Viktor},
month = nov,
year = 2021,
publisher = {GitHub},
journal = {GitHub repository},
howpublished = {\url{https://github.com/Denys88/rl_games}},
}

@article{iHY-Hand,
author = {Lael U. Odhner and Leif P. Jentoft and Mark R. Claffee and Nicholas Corson and Yaroslav Tenzer and Raymond R. Ma and Martin Buehler and Robert Kohout and Robert D. Howe and Aaron M. Dollar},
title ={A compliant, underactuated hand for robust manipulation},
journal = {The International Journal of Robotics Research},
volume = {33},
number = {5},
pages = {736-752},
year = {2014},
doi = {10.1177/0278364913514466},
URL = { https://doi.org/10.1177/0278364913514466},
}

@INPROCEEDINGS{Eagle_Shoal,
  author={Wang, Tao and Geng, Zhanxiao and Kang, Bo and Luo, Xiaochuan},
  booktitle={2019 International Conference on Robotics and Automation (ICRA)}, 
  title={Eagle Shoal: A new designed modular tactile sensing dexterous hand for domestic service robots}, 
  year={2019},
  volume={},
  number={},
  pages={9087-9093},
  doi={10.1109/ICRA.2019.8793842}
}

@UNPUBLISHED{Makoviychuk2021-ll,
  title    = "Isaac Gym: High Performance {GPU} Based Physics Simulation For
              Robot Learning",
  author   = "Makoviychuk, Viktor and Wawrzyniak, Lukasz and Guo, Yunrong and
              Lu, Michelle and Storey, Kier and Macklin, Miles and Hoeller,
              David and Rudin, Nikita and Allshire, Arthur and Handa, Ankur and
              {Gavriel State}",
  journal  = "https://openreview.net › forumhttps://openreview.net › forum",
  month    =  nov,
  year     =  2021
}

@article{shaw2023leap,
      title={LEAP Hand: Low-Cost, Efficient, and Anthropomorphic Hand for Robot Learning}, 
      author={Kenneth Shaw and Ananye Agarwal and Deepak Pathak},
      journal={arXiv preprint arXiv:2309.06440},
      year={2023},
      archivePrefix={arXiv},
      primaryClass={cs.RO}
}

@article{toshimitsu2023getting,
      title={Getting the Ball Rolling: Learning a Dexterous Policy for a Biomimetic Tendon-Driven Hand with Rolling Contact Joints}, 
      author={Yasunori Toshimitsu and Benedek Forrai and Barnabas Gavin Cangan and Ulrich Steger and Manuel Knecht and Stefan Weirich and Robert K. Katzschmann},
      journal={2023 IEEE-RAS 22\textsuperscript{nd} International Conference on Humanoid Robots (Humanoids)},
      pages={1--7},
      year={2023},
      organization={IEEE}
}

@INPROCEEDINGS{YCBcomplete,
  author={Calli, Berk and Singh, Arjun and Walsman, Aaron and Srinivasa, Siddhartha and Abbeel, Pieter and Dollar, Aaron M.},
  booktitle={2015 International Conference on Advanced Robotics (ICAR)}, 
  title={The YCB object and Model set: Towards common benchmarks for manipulation research}, 
  year={2015},
  volume={},
  number={},
  pages={510-517},
  doi={10.1109/ICAR.2015.7251504}
}

@inproceedings{lugaresi2019mediapipe,
  title={Mediapipe: A framework for perceiving and processing reality},
  author={Lugaresi, Camillo and Tang, Jiuqiang and Nash, Hadon and McClanahan, Chris and Uboweja, Esha and Hays, Michael and Zhang, Fan and Chang, Chuo-Ling and Yong, Ming and Lee, Juhyun and others},
  booktitle={Third workshop on computer vision for AR/VR at IEEE computer vision and pattern recognition (CVPR)},
  volume={2019},
  year={2019}
}

@article{cutkosky1989grasp,
  title={On grasp choice, grasp models, and the design of hands for manufacturing tasks.},
  author={Cutkosky, Mark R and others},
  journal={IEEE Transactions on robotics and automation},
  volume={5},
  number={3},
  pages={269--279},
  year={1989}
}

@article{rahman2013design,
author = {Akhlaquor Rahman and Adel Al-Jumaily},
title ={Design and Development of a Bilateral Therapeutic Hand Device for Stroke Rehabilitation},
journal = {International Journal of Advanced Robotic Systems},
volume = {10},
number = {12},
pages = {405},
year = {2013},
doi = {10.5772/56809},
URL = {    
        https://doi.org/10.5772/56809
},
}

@inproceedings{samadikhoshkho2019brief,
  title={A brief review on robotic grippers classifications},
  author={Samadikhoshkho, Zahra and Zareinia, Kourosh and Janabi-Sharifi, Farrokh},
  booktitle={2019 IEEE Canadian Conference of Electrical and Computer Engineering (CCECE)},
  pages={1--4},
  year={2019},
  organization={IEEE}
}

@inproceedings{sadun2016grasping,
  title={Grasping analysis for a 3-finger adaptive robot gripper},
  author={Sadun, Amirul Syafiq and Jalani, Jamaludin and Jamil, Faizal},
  booktitle={2016 2\textsuperscript{nd} IEEE International Symposium on Robotics and Manufacturing Automation (ROMA)},
  pages={1--6},
  year={2016},
  organization={IEEE}
}

@article{terrile2021comparison,
  title={Comparison of different technologies for soft robotics grippers},
  author={Terrile, Silvia and Arg{\"u}elles, Miguel and Barrientos, Antonio},
  journal={Sensors},
  volume={21},
  number={9},
  pages={3253},
  year={2021},
  publisher={MDPI}
}

@article{andrychowicz2020learning,
  title={Learning dexterous in-hand manipulation},
  author={Andrychowicz, OpenAI: Marcin and Baker, Bowen and Chociej, Maciek and Jozefowicz, Rafal and McGrew, Bob and Pachocki, Jakub and Petron, Arthur and Plappert, Matthias and Powell, Glenn and Ray, Alex and others},
  journal={The International Journal of Robotics Research},
  volume={39},
  number={1},
  pages={3--20},
  year={2020},
  publisher={SAGE Publications Sage UK: London, England}
}

@article{schulman2017proximal,
  title={Proximal policy optimization algorithms},
  author={Schulman, John and Wolski, Filip and Dhariwal, Prafulla and Radford, Alec and Klimov, Oleg},
  journal={arXiv preprint arXiv:1707.06347},
  year={2017}
}

@inproceedings{appius2022raptor,
  title={Raptor: Rapid aerial pickup and transport of objects by robots},
  author={Appius, Aurel X and Bauer, Erik and Bl{\"o}chlinger, Marc and Kalra, Aashi and Oberson, Robin and Raayatsanati, Arman and Strauch, Pascal and Suresh, Sarath and von Salis, Marco and Katzschmann, Robert K},
  booktitle={2022 IEEE/RSJ International Conference on Intelligent Robots and Systems (IROS)},
  pages={349--355},
  year={2022},
  organization={IEEE}
}

@inproceedings{zlokapa2022integrated,
  title={An integrated design pipeline for tactile sensing robotic manipulators},
  author={Zlokapa, Lara and Luo, Yiyue and Xu, Jie and Foshey, Michael and Wu, Kui and Agrawal, Pulkit and Matusik, Wojciech},
  booktitle={2022 International Conference on Robotics and Automation (ICRA)},
  pages={3136--3142},
  year={2022},
  organization={IEEE}
}

@article{piazza_century_2019,
	title = {A Century of Robotic Hands},
	volume = {2},
	issn = {2573-5144, 2573-5144},
	url = {https://www.annualreviews.org/doi/10.1146/annurev-control-060117-105003},
	doi = {10.1146/annurev-control-060117-105003},
	pages = {1--32},
	number = {1},
	journaltitle = {Annual Review of Control, Robotics, and Autonomous Systems},
	shortjournal = {Annu. Rev. Control Robot. Auton. Syst.},
	author = {Piazza, C. and Grioli, G. and Catalano, M.G. and Bicchi, A.},
	urldate = {2024-09-06},
	date = {2019-05-03},
	langid = {english},
}

\end{document}